\documentclass[journal]{IEEEtran}

\usepackage[T1]{fontenc}
\usepackage{amsmath,amssymb,amsfonts}
\usepackage[utf8]{inputenc} 
\usepackage{graphicx}
\usepackage{amsfonts}
\usepackage{mathtools}
\usepackage{siunitx}
\usepackage{tabularx}
\usepackage{booktabs}
\usepackage{bm}
\usepackage[nolist,nohyperlinks]{acronym}

\usepackage{tikz}
\usetikzlibrary{matrix}

\usepackage{url}

\usepackage{svg}
\usepackage[textsize=scriptsize,color=blue!20!white]{todonotes}

\usepackage[table]{xcolor}

\newcommand{\vc}[1]{\ensuremath{\bm{#1}}}

\newcommand{\bmat}[1]{\ensuremath{\begin{bmatrix}#1\end{bmatrix}}}

\DeclarePairedDelimiter\abs{\lvert}{\rvert}
\DeclarePairedDelimiter\norm{\lVert}{\rVert}
\makeatletter
\let\oldabs\abs
\let\oldnorm\norm
\def\abs{\@ifstar{\oldabs}{\oldabs*}}
\def\norm{\@ifstar{\oldnorm}{\oldnorm*}}
\makeatother

\DeclareMathOperator{\trace}{tr}

\def\-{\raisebox{0pt}{-}}

\newcommand{\argmin}[1]{\underset{#1}{\operatorname{arg}\,\operatorname{min}}\;}

\newcommand{\SE}[1]{\ensuremath{\mathrm{SE}({#1})}}

\newcommand{\tr}[1]{\trace{\left(#1\right)}}

\let\oldunderbrace\underbrace
\renewcommand{\underbrace}[2]{\let\scriptstyle\textstyle \oldunderbrace{#1}_{#2}}
\definecolor{tab10blue}{rgb}{0.12156863, 0.46666667, 0.70588235}

\DeclareFontFamily{U}{mathx}{\hyphenchar\font45}
\DeclareFontShape{U}{mathx}{m}{n}{ <-> mathx10 }{}
\DeclareSymbolFont{mathx}{U}{mathx}{m}{n}
\DeclareFontSubstitution{U}{mathx}{m}{n}
\DeclareMathAccent{\widebar}{\mathalpha}{mathx}{"73}
\newcolumntype{C}[1]{>{\centering\let\newline\\\arraybackslash\hspace{0pt}}m{#1}}
\newcolumntype{D}[1]{>{\centering\let\newline\\\arraybackslash\hspace{0pt}\columncolor{gray!20}}m{#1}}
\newcolumntype{R}[1]{>{\raggedleft\let\newline\\\arraybackslash\hspace{0pt}}m{#1}}
\newcolumntype{L}[1]{>{\raggedright\let\newline\\\arraybackslash\hspace{0pt}}m{#1}}
\newcolumntype{M}[1]{>{\raggedright\let\newline\\\arraybackslash\hspace{0pt}}m{#1}}
\newcolumntype{N}{>{\refstepcounter{rowcntr}\therowcntr}c}

\usepackage{tabstackengine}
\stackMath
\setstackgap{L}{18pt}
\setstacktabbedgap{0pt}
\def\lrgap{\kern6pt}

\usepackage{pifont}
\newtheorem{theorem}{Theorem}
\acrodef{ADMM}[ADMM]{alternating direction method of multipliers}
\acrodef{dSDP}[dSDP]{decomposed SDP}
\acrodef{EVR}[EVR]{Eigenvalue ratio}
\acrodef{iSAM}[iSAM]{incremental smoothing and mapping}
\acrodef{GP}[GP]{Gaussian process}
\acrodef{GN}[GN]{Gauss-Newton}
\acrodef{mae}[MAE]{mean absolute error}
\acrodef{NLS}[NLS]{nonlinear least squares}
\acrodef{PSD}[PSD]{positive semidefinite}
\acrodef{QCQP}[QCQP]{quadratically-constrained quadratic program}
\acrodef{SDP}[SDP]{semidefinite program}
\acrodef{SLAM}[SLAM]{simultaneous localization and mapping}
\acrodef{STD}[STD]{standard deviation}
\acrodef{CTRO}[CT-RO]{continuous-time range-only localization}
\acrodef{MW}[MW]{matrix-weighted localization}

\newcommand\citet[1]{\cite{#1}}

\usepackage[hidelinks]{hyperref} 

\begin{document}

\title{Exploiting Chordal Sparsity for\\ Globally Optimal Estimation with Factor Graphs}

\author{Avinash Subramanian$^{1}$, Connor Holmes$^{2}$, Timothy D. Barfoot$^{2}$, Frank Dellaert$^{1}$, and Frederike Dümbgen$^{3}$
\thanks{$^{1}$Avinash Subramanian and Frank Dellaert are with the College of Computing, Georgia Institute of Technology, Atlanta, GA, USA,
        {\tt\small asubramanian97@gatech.edu, frank.dellaert@cc.gatech.edu}.}%
\thanks{$^{2}$ Connor Holmes and Timothy Barfoot are with the Robotics Institute, University of Toronto, Toronto, ON, Canada, {\tt\small tim.barfoot@utoronto.ca, connor.holmes@utoronto.ca}.}
\thanks{$^{3}$Frederike Dümbgen is with the Department of Mechanical Engineering, Carnegie Mellon University, Pittsburgh, PA, USA,
        {\tt\small fduembgen@cmu.edu}.}
        }

\markboth{ICRA 2026 Workshop on Frontiers of Optimization for Robotics}%
{Shell \MakeLowercase{\textit{et al.}}: Bare Demo of IEEEtran.cls for IEEE Journals}

\maketitle

\begin{abstract}
Robust and efficient state estimation is crucial for perception, navigation, and control in robotics. State estimation problems are conveniently modeled using the factor-graph framework as enabled by modern software packages such as \textit{GTSAM} or \textit{g2o}. However, the standard solvers included in such frameworks are local and may converge to poor local minima, posing significant safety concerns. Conversely, techniques based on convex relaxations have been shown to provide a means of globally solving or certifying many state estimation problems. However, these relaxations 1) often require substantial effort to formulate, and 2) may incur significantly higher cost compared to efficient local solvers, as they require solving a large semidefinite program (SDP). 
In this work, we address both shortcomings by 1) creating a new procedure within the \textit{GTSAM} framework for automatically constructing convex SDP relaxations for any factor graphs with common factor and variable types, and by 2) exploiting the Bayes tree constructions native to \textit{GTSAM} to decompose the SDP problem, leading to significant speedup in solver time for chordally sparse problems. We demonstrate the favorable scaling of this structure-exploiting \textit{global} estimator compared to standard local solvers for two case studies: A 3D pose-graph SLAM problem with a ring factor graph and a 2D localization problem with a chain factor graph. The software framework is available at \url{https://github.com/borglab/gtsam}.
\end{abstract}


\IEEEpeerreviewmaketitle

\acresetall

\section{Introduction}

Fast and accurate state estimation is crucial for autonomous navigation and control of robots~\cite{barfoot_state_2017}. Factor graphs have emerged as a powerful framework, popular amongst robotics practitioners, for modeling complex estimation and sensor fusion problems: They are defined as a bipartite graph with edges between variable nodes (denoting states to be estimated) and factor nodes (denoting various measurements or priors)~\cite{dellaert_factor_2021}. The factor graph modeling framework has been democratized by state-of-the-art software implementations such as \textit{GTSAM} ~\cite{dellaert2012factor} and \textit{g2o}~\cite{kummerle2011g}: These provide a comprehensive set of pose parameterizations, measurement and prior factor types, as well as efficient local optimization solvers such as Gauss-Newton and Levenberg-Marquardt for solving the estimation problem. While these local solvers are computationally efficient, they typically require high-quality initialization and do not provide guarantees of global optimality, which render them risky for safety-critical applications.


Recently, certifiably optimal state estimation techniques~\cite{carlone_initialization_2015} have been proposed that leverage the fact that polynomial optimization problems often admit a tight convex relaxation in the form of an \ac{SDP}~\cite{lasserre_convergent_2006,parrilo_semidefinite_2003}. These methods provide \textit{a posteriori} optimality guarantees and scale favorably compared to \textit{always-optimal}, but worst-case exponential, branch-and-bound methods. However, the naive use of~\ac{SDP} solvers for the resulting relaxations may incur prohibitive costs.
To address this issue, a number of related works have proposed techniques to exploit problem structure in the quest for a scalable certifiably global state estimator. Rosen et al.~\cite{rosen_se-sync_2019} introduced SE-Sync, which exploits the low-rank structure of SDP solutions through the Burer-Monteiro factorization~\cite{burer_local_2005} by applying the Riemannian staircase method. Tian et al.~\cite{tian_distributed_2021} extended this work to use a distributed algorithm via Riemannian block coordinate descent.

\begin{figure*}[t]
    \centering
    \includegraphics[width=\textwidth, trim={0 0em 0 0}, clip]{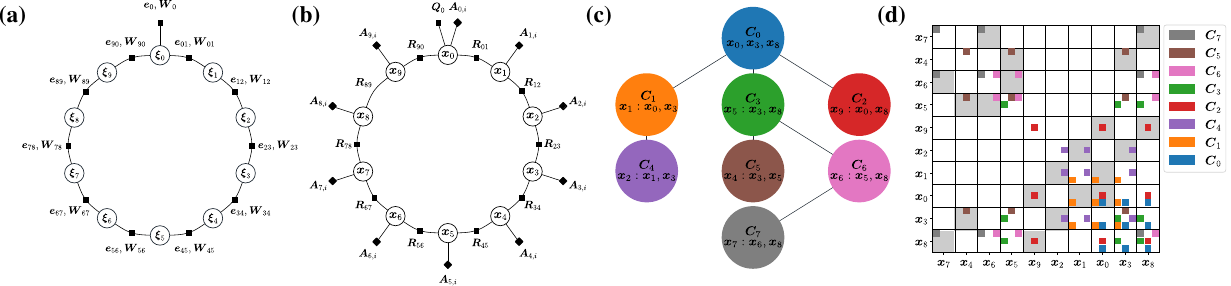}
    
  \vspace{-1em}
    \caption{Visualization of the proposed framework. (a) The user creates the factor graph representation of the problem (here, a 3D pose-graph SLAM problem with 10 variables is shown), (b) we automatically lift the factor graph to a \acf{QCQP} in lifted variables $\bm{x}_i$, (c) we construct the Bayes tree through variable elimination, which is then used to directly create a clique-based \acf{SDP} problem (d): instead of solving for one large \acf{PSD} variable, we solve for smaller \ac{PSD} variables (one per clique), shown by the corresponding colors.}
    \label{fig:bayes_tree}
\end{figure*}

Other works have focused on exploiting the chain-like sparsity structure exhibited by a subset of estimation problems. Building on the seminal works from the \ac{SDP} literature~\cite{vandenberghe_chordal_2015, zheng_chordal_2021}, Duembgen et al.~\cite{dumbgen2024exploiting} introduced a distributed global estimator that can exploit chordal sparsity, and laid some foundations in using the factor-graph language to simplify problem formulation. However, the empirical validation and provided software were limited to specific problems and sparsity patterns, \emph{i.e.}, localization with chain-like sparsity. Going beyond the (first-order) Shor's relaxation, Kang et al. ~\cite{kang_fast_2024} showcased a GPU-accelerated solver for trajectory optimization exploiting chain-like sparsity also for higher-order moment-hierarchy relaxations. Graesdal et al. \cite{graesdal2024towards} proposed a method for contact-rich motion planning where the same chain-like sparsity structure is exploited in the setting of graphs-of-convex-sets relaxations. 

While some software packages that exploit chordal sparsity exist~\cite{Garstka_2021,sparse-Lasserre}, they come with software interfaces not necessarily suited for roboticists and do not always allow for control of the (non-unique) decomposition technique used. The aforementioned use cases in robotics instead manually decompose the chain-like problems, but this approach does not scale well to more complex sparsity patterns. Instead, we propose integrating chordal decomposition techniques into \textit{GTSAM}, which is commonly used to formulate robotics problems, and is already equipped with variable elimination and chordal decomposition tools. In concurrent work, the fast \ac{SDP} solver techniques based on the Riemannian staircases are also being integrated in \textit{GTSAM}~\cite{xu2026certifiable}, which will significantly reduce the effort required for practitioners to investigate the efficacy of alternative structure-exploiting solvers for their use cases.
In summary, we make the following contributions:
\begin{enumerate}
    \item We develop an easy-to-use \textit{GTSAM}-based software framework that democratizes access to \ac{SDP}-based certifiable estimators.
    \item We show how to use variable elimination on a factor graph, and the resulting Bayes tree, to create a chordally decomposed estimator for arbitrary sparsity patterns beyond commonly treated chains.
    \item We demonstrate the favorable scaling of this estimator on two problems of interest:  A 3D pose-graph SLAM problem modeled as a ring factor graph with poses in \SE{3}, and a 2D localization problem modeled using a chain factor graph with poses in \SE{2}.
\end{enumerate}
\section{Method}
\label{sec:methodology}
In this section, we show how to obtain a computationally efficient convex relaxation for many canonical state estimation problems by leveraging both the graphical modeling capabilities and efficient factorization procedures provided by \textit{GTSAM}. The workflow is visualized in Figure~\ref{fig:bayes_tree} for the 3D pose-graph SLAM problem with a ring factor-graph structure.
In particular, we first construct the original estimation problems using a standard factor graph (Section~\ref{sec:fg}), then lift the problem to an equivalent Quadratically Constrained Quadratic Program (QCQP) (Section~\ref{sec:qcqp}), and finally perform variable elimination to construct a Bayes tree (Section~\ref{sec:bayes}). The cliques of the Bayes tree correspond to the chordally decomposed \ac{SDP} relaxation. Importantly, the entire process following the factor graph setup can be automated, resulting in an easy-to-use yet certifiably optimal estimation pipeline. 
\label{sec:fg}
Most common state estimation problems can be formulated as inference on a factor graph.
State estimation involves determining the maximum a-posteriori estimate of the robot state from noisy measurements,  equivalent to solving a nonlinear least squares optimization problem assuming zero-mean Gaussian noise and conditional independence: 
\begin{equation}
  \begin{aligned}
    \{\hat{\vc{\xi}_i}\}_{i=1}^{N} = \argmin{\vc{\xi}_i \in \mathcal{X}, i\in[N]}& \sum_{k \in \mathcal{Q}} \vc{e}_{k}(\vc{\xi}_k)^\top\vc{W}_{k}\vc{e}_{k}(\vc{\xi}_k) \\
    + \sum_{(i,j)\in\mathcal{R}} &\vc{e}_{ij}(\vc{\xi}_i, \vc{\xi}_j)^\top \vc{W}_{ij}\vc{e}_{ij}(\vc{\xi}_i, \vc{\xi}_j),
  \end{aligned}
  \label{eq:nls}
\end{equation}
where $\vc{\xi}_i$, $i\in[N]:=\{1, \ldots, N\}$ represent the state variables to be estimated, with $\mathcal{X}$ often chosen to be $\mathrm{SO}(d)$ or $\mathrm{SE}(d)$. We denote by $\mathcal{Q}$ ($\mathcal{R}$) the set of indices (pairs) at which we have absolute (relative) measurements or prior terms. The cost terms of the first and second sum correspond to unary (absolute) factors and binary (odometry) factors, respectively.\footnote{Note that we focus on unary and binary factors, which are the most common and the only types required for the presented examples. The framework does allow for any factor types.} These are defined by their respective error terms $\vc{e}_{k}(\cdot)$, $\vc{e}_{ij}(\cdot)$ and corresponding inverse covariance matrices $\vc{W}_{k}$, $\vc{W}_{ij}$. Problem~\eqref{eq:nls} is a nonconvex optimization problem for nonlinear $\vc{e}_k$, $\vc{e}_{ij}$ or for a nonconvex set $\mathcal{X}$. Problem~\eqref{eq:nls} can be represented by a factor graph $\mathcal{G}_f=(\mathcal{V},\mathcal{M},\mathcal{E})$~\cite{dellaert_factor_2021}, where $\mathcal{V}$ contains all states, $\mathcal{M}$ contains the measurements or priors, and $\mathcal{E}$ defines the (bipartite) connectivity between $\mathcal{V}$ and $\mathcal{M}$.
\vspace{1mm}

\begin{figure*}[t]
  \centering
\makebox[\textwidth][c]{%
  \includegraphics[height=4.8cm]{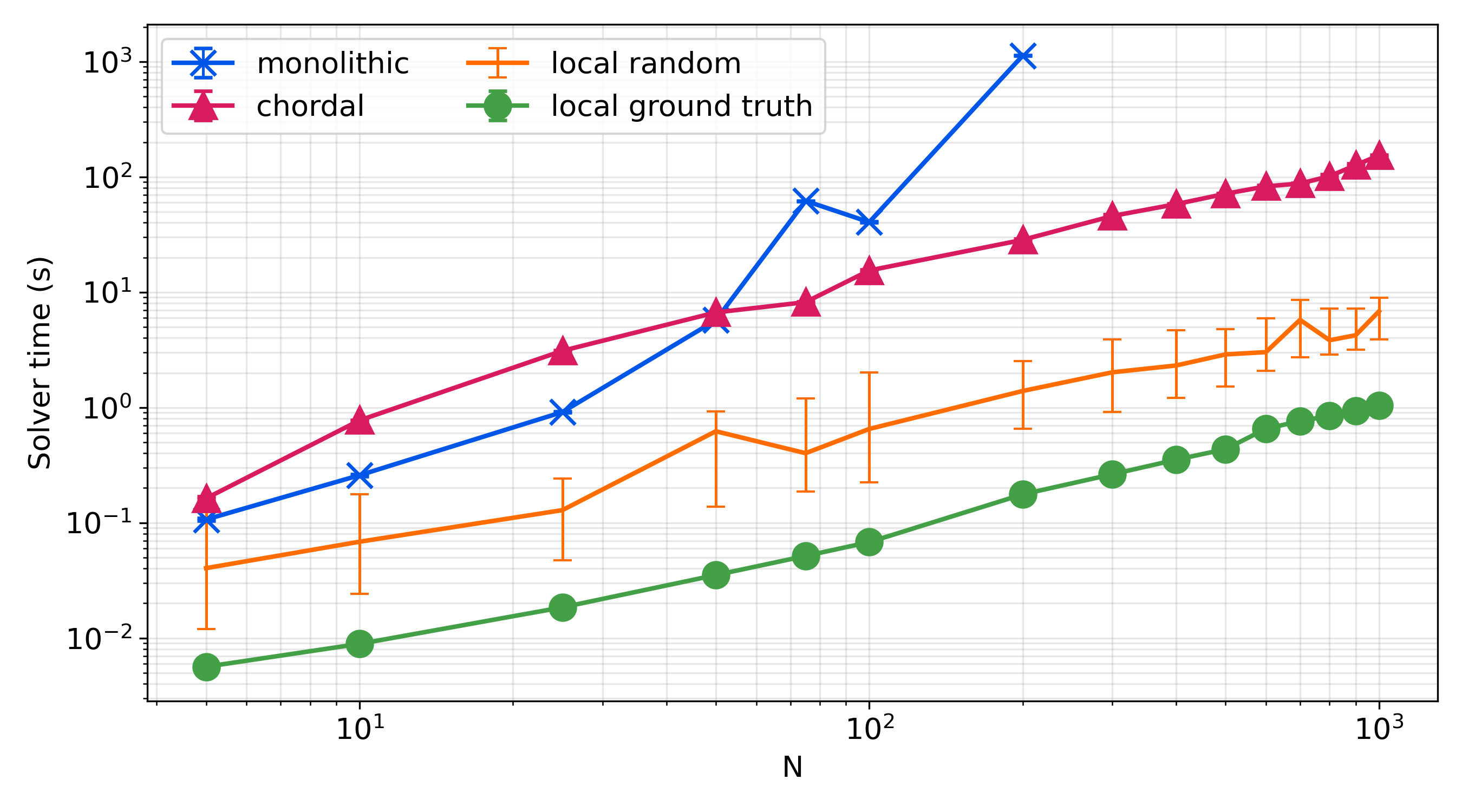}
  \hfill
  \includegraphics[height=4.8cm]{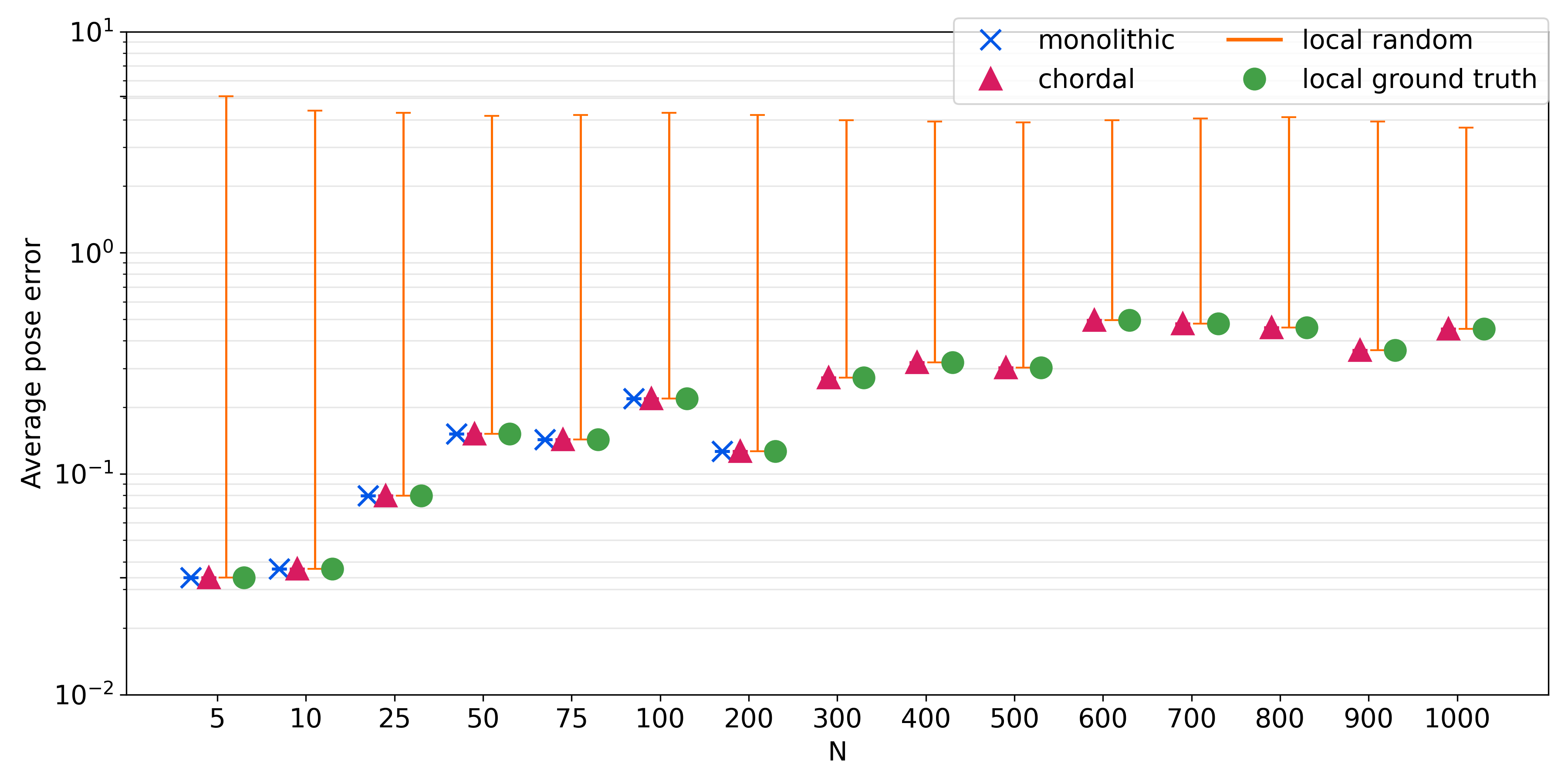}
  }
  
  \vspace{-1em}
  \caption{Solver times (left) and Average pose error (right) for the 3D pose-graph SLAM problem with a ring factor graph as a function of the number of poses $N$. The presented \textit{chordal} estimator has a complexity comparable with local solvers and with little variance in solution time and accuracy.}
  \label{fig:se3-ring-scalability}
\end{figure*}
\subsubsection*{Example --- 3D SLAM with ring factor graph} Figure~\ref{fig:bayes_tree}(a) presents the ring factor graph for an example 3D pose-graph SLAM problem. The variable nodes of the factor graph are the poses $\vc{\xi}_i: \vc{T}_{i0}\in\SE{3}$ to be estimated. The odometry (binary) factors are noisy relative pose measurements $\vc{e}_{ij} = \|\vc{T}_{i0}-\widetilde{\vc{T}}_{ij} \vc{T}_{j0}\|^2_{\sigma_{ij}}$ where $\widetilde{\vc{T}}_{ij}$ is a measurement of the relative transform from pose $i$ to pose $j$ and $\widetilde{\vc{T}}_{i0}$ a measurement of pose $i$. We are also given an absolute measurement on the initial pose, which forms a unary factor $\vc{e}_{k} = \|\vc{T}_{i0}-\widetilde{\vc{T}}_{i0}\|^2_{\sigma_{i}}$). We assume isotropic noise for all measurements thus $\vc{W}_{ij}=\sigma_{ij}^{-2} \vc{I}$ and $\vc{W}_{k} = \sigma_{i}^{-2} \vc{I}$.
\subsection{Lifting to a QCQP}
\label{sec:qcqp}
The variables and factors of Problem~\eqref{eq:nls} are lifted to a higher-dimensional space to yield a \ac{QCQP} for which standard SDP convex relaxations exist \cite{lasserre2015introduction, nie2023moment}. Importantly, this operation leaves the graph's original connectivity unchanged. The intuition behind variable lifting is to introduce enough higher-order terms such that the cost and the constraints stemming from $\mathcal{X}$ in Problem~\eqref{eq:nls} become quadratic in the lifted vectors. The state $\vc{\xi}_i$ yields the lifted variable
\begin{equation}
  \vc{x}_i \in \mathbb{R}^D, \vc{x}_i := \bmat{1 & \vc{\xi}_i^\top & \vc{\ell}_i(\vc{\xi}_i)^\top}^\top, 
  \label{eq:xik}
\end{equation}
where we have introduced the lifting function $\vc{\ell}_i$ and the homogenization constant 1, which allows for constant and linear terms to be represented using quadratic terms as well. 


The QCQP reformulation of~\eqref{eq:nls} takes the form:
\begin{equation}
  \begin{aligned}
    \{\hat{\vc{x}}_k\}_{k\in\mathcal{V}} = &\argmin{\vc{x}_k, k \in \mathcal{V}} \sum_{k\in\mathcal{Q}} \vc{x}_k^\top\vc{Q}_k\vc{x}_k + \sum_{(i, j)\in\mathcal{R}}\vc{x}_{ij}^\top\vc{R}_{ij}\vc{x}_{ij} \\
    \text{s.t. }
    &\vc{x}_k^\top \vc{A}_{j_k}^s \vc{x}_k = b_{j_k}, \quad k\in\mathcal{V},j_k\in[N_{k}^s] \\
    &\vc{x}_k^\top \vc{A}_{j_k}^p \vc{x}_k = 0, \quad k\in\mathcal{V},\,j_k\in[N_{k}^p] 
  \end{aligned},
  \label{eq:qcqp-split}
\end{equation}
where $\vc{x}_{ij}$ is the concatenation of $\vc{x}_i$ and $\vc{x}_j$.
The constraint matrices $\vc{A}_{j_k}^s$ and scalars $b_{j_k}$, $j_k\in[N_k^s]$ enforce the substitution constraints from~\eqref{eq:xik} for lifted variable $k$, and $\vc{A}_{j_k}^p$, $j_k\in[N_k^p]$ enforce the primary constraints from $\mathcal{X}$. 
The cost matrices $\vc{Q}_k$ and $\vc{R}_{ij}$ are derived in closed form for each of the measurement and prior types.
This \ac{QCQP} can be represented as a \textit{lifted} factor graph as shown in Figure~\ref{fig:bayes_tree}(b).



\subsection{Bayes Tree Construction}\label{sec:bayes}

By collecting all variables in one large vector $\bm{x}$, the \ac{QCQP} could be relaxed to an \ac{SDP} with variable $\bm{X}\succeq 0$ of size $ND$, obtained by forming the rank-one variable $\bm{X}=\bm{x}\bm{x}^\top$ and relaxing its rank constraint. For the detailed reformulation, we refer the reader to~\cite{dumbgen2024exploiting}. However, the cost to solve such an \ac{SDP} would in general be approximately cubic in its dimension. We will refer to this formulation as the \textit{monolithic}~\ac{SDP}. 

Instead, we formulate an equivalent~\ac{SDP} problem with multiple, often significantly smaller, \ac{PSD} variables. Starting from the QCQP factor graph, a variable elimination (factorization) procedure is executed using a specific ordering (\emph{e.g.}, COLAMD or METIS) as explained in detail by Dellaert and Kaess~\cite[Ch.~4]{dellaert2017factor}. This yields a Bayes net, which is known to always be chordal~\cite{dellaert2017factor}. The maximal cliques of the Bayes net can be listed in an undirected tree called a clique tree (also known as a junction tree) or a directed tree that preserves the order of eliminations called a Bayes tree~\cite{dellaert2017factor}. The Bayes tree is defined as $\vc{T}_c:={(\mathcal{V}_c,\mathcal{E}_c)}$, where the vertices index the maximal cliques, and edges between two cliques indicate shared variables. Figure~\ref{fig:bayes_tree}(c) illustrates the Bayes tree attained by performing variable elimination on the shown QCQP factor graph with a METIS ordering. 

By associating $T$ smaller PSD matrices $\vc{C}_t$, $t\in\mathcal{V}_c$ with the maximal cliques obtained from the Bayes tree, we arrive at a chordally decomposed SDP formulation:
\begin{equation}
  \begin{aligned}
    \{\hat{\vc{C}}_i\}_{t\in\mathcal{V}_c} = \argmin{\vc{C}_1, \ldots, \vc{C}_T} & \sum_{t\in\mathcal{V}_c} \tr{\vc{Q}_t\vc{C}_t} \\
    \text{s.t. } & \mathcal{A}_t(\vc{C}_t)= \vc{b}_t, \quad t\in\mathcal{V}_c \\
    & \mathcal{S}_t^{t'}(\vc{C}_t)=\mathcal{S}_{t'}^t(\vc{C}_{t'}), \quad t,t' \in \mathcal{E}_c \\
    & \vc{C}_t\succeq 0, \quad t\in\mathcal{V}_c
  \label{eq:dsdp}
  \end{aligned},
\end{equation}
where $\mathcal{A}_t$ is the linear operator stacking all primary, substitution, and possibly redundant constraints\footnote{Redundant constraints are irrelevant to the original problem but yield a tighter convex relaxation (closer to the global optimum) without changing the original feasible set~\cite{dumbgen_toward_2024}. We implicitly assume that redundant constraints do not change the problem sparsity. If redundant constraints add novel connections in the original factor graph, they must be added \textit{before} the Bayes tree construction.} of clique $t$ , and $\mathcal{S}_t^{t'}$ is the operator selecting the shared variables with $t'$ from clique $t$. $\vc{Q}_t$ is the clique-wise cost matrix, obtained from splitting the cost variables equally across overlapping nodes. The decomposition of the \ac{PSD} variable into its cliques is shown in Figure~\ref{fig:bayes_tree}(d).


We note that the fact that the chordally decomposed SDP Problem~\ref{eq:dsdp} is equivalent to the \textit{monolithic}~\ac{SDP} is a consequence of the fundamental result by Fukuda et al. \cite{fukuda_exploiting_2001}: If the aggregate sparsity pattern of cost and constraint matrices corresponds to a chordal graph, the large \ac{PSD} constraint $\vc{X}\succeq 0$ can be decomposed into a set of smaller PSD constraints on the maximal cliques of the graph, coupled by equality constraints on their intersections~\cite{garstka_clique_2020}. The key insight leveraged in our formulation is that the cliques of the Bayes tree correspond to the maximal cliques of the chordal completion of the sparsity pattern of the \textit{monolithic} SDP.

\begin{figure*}[t]
  \centering
    \makebox[\textwidth][c]{%
  \includegraphics[height=4.8cm]{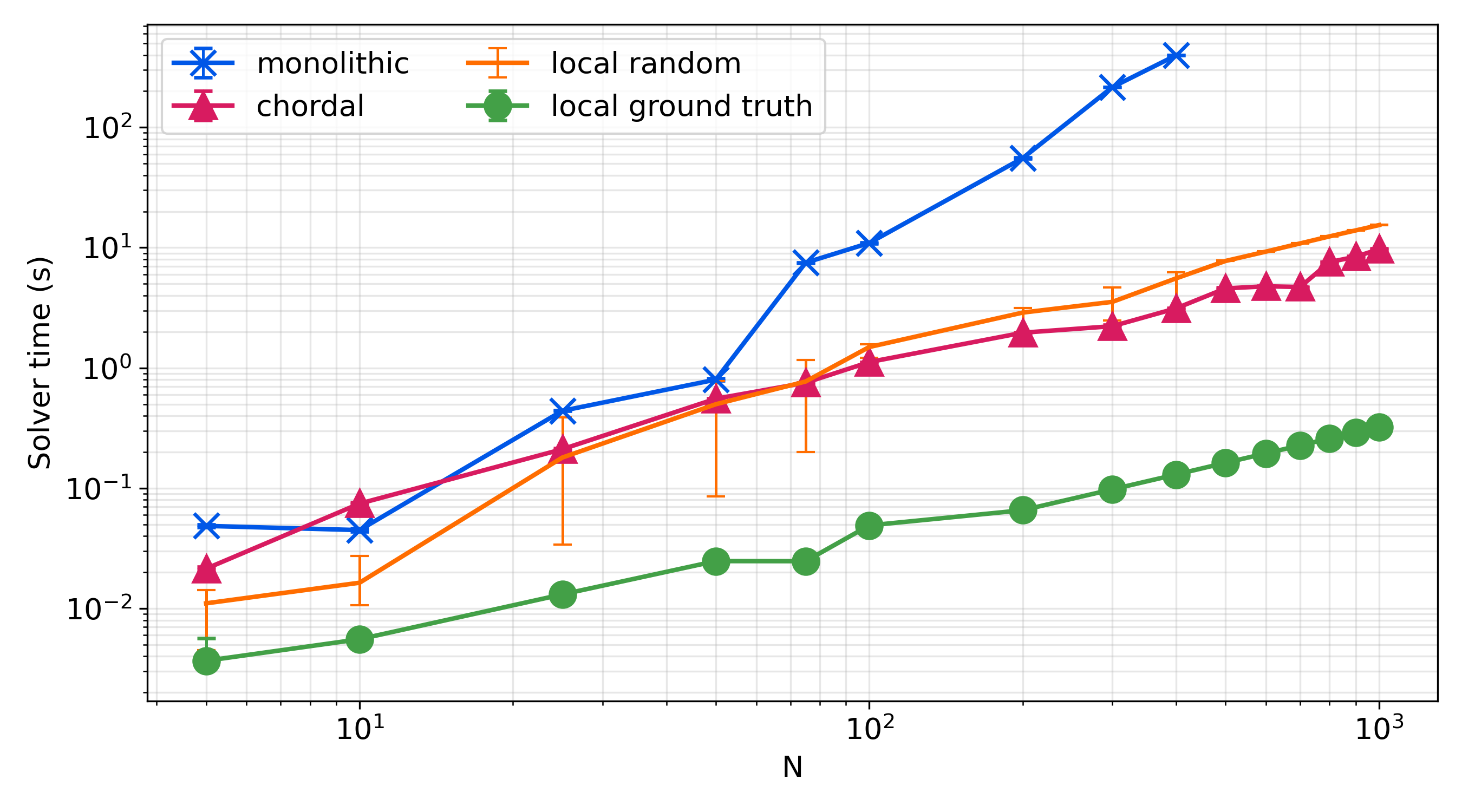}
  \includegraphics[height=4.8cm]{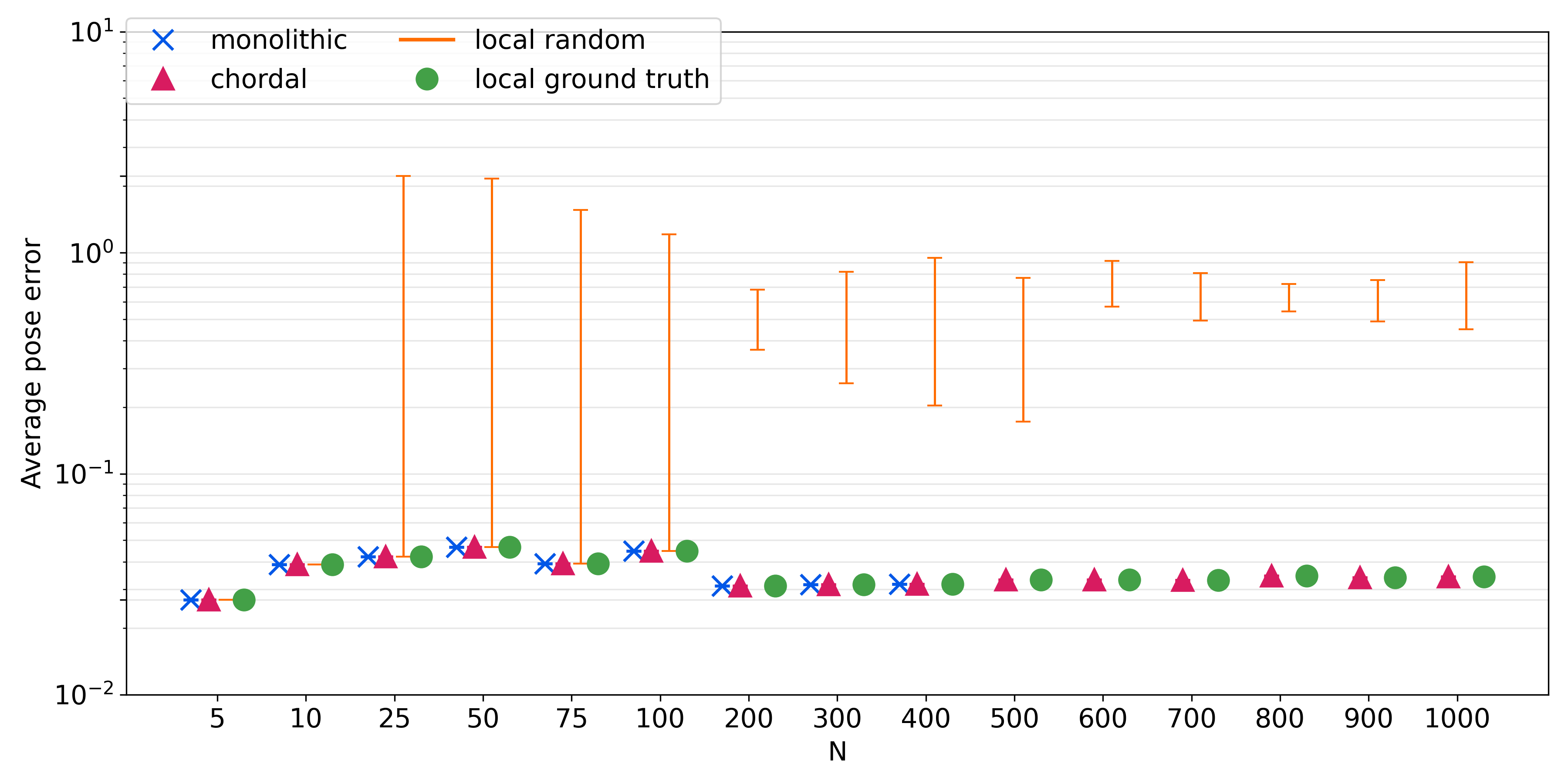}
  }
  
  \vspace{-1em}
  \caption{Solver times (left) and Average pose error (right) for the 2D pose localization problem with a chain factor graph representation as a function of the number of poses $N$. The presented \textit{chordal} estimator has complexity comparable to that of local solvers, with little variance in solution time and accuracy.}
  \label{fig:se2-scalability}
\end{figure*}

\subsection{Implementation}
The lifting procedure and the formulation of the SDP problem are implemented in C++ using the MOSEK fusion API for C++ \cite{aps2019mosek}. Similar to GTSAM's implementation, a template metaprogramming approach is used to generate specialized code at compile time for the different matrix Lie group types (currently \texttt{Rot2}, \texttt{Rot3}, \texttt{Pose2}, \texttt{Pose3} supported) and factor types (currently \texttt{FrobeniusBetween}, \texttt{FrobeniusPrior}, \texttt{PriorFactor} supported) of \textit{GTSAM}.\footnote{Note that any factor and variable types that can be encoded with polynomial cost and constraints are suitable for the lifting procedure.}


\section{Results and Discussion}

The proposed method is applied to a 3D pose-based SLAM problem represented by a ring factor graph and a 2D localization problem represented by a chain factor graph. For comparison of solver times and accuracy of estimated poses, three other estimation algorithms are implemented: The \textit{monolithic} SDP estimator that does not exploit any sparsity structure but results in a certifiably global estimate, \textit{local random} and \textit{local ground truth}, which use \textit{GTSAM}'s Levenberg-Marquardt nonlinear least-squares solver with starting values chosen as random poses or ground truth poses, respectively. All experiments are run on an Apple M1 MacBook Pro.


\subsection{3D pose-graph SLAM example}
\label{sec:3D-results}
We first treat the 3D pose-graph SLAM problem with ring structure described in Section \ref{sec:fg}. We study the scalability of the proposed chordally sparse SDP formulation as the number of poses $N$ increases. All problems have the same fixed scale; thus, increasing $N$ increases the sampling density of measurements. For each value of $N$, the sensor noise is randomly sampled and composed with the nominal measurement. Each of the 4 estimators is run five times with the same measurement, such that the only stochasticity arises from randomness in the estimation algorithm (in starting value or convergence). The empirical results in the left plot of Figure~\ref{fig:se3-ring-scalability} show that the solution time of the \textit{chordal} estimator grows (roughly) linearly in the number of poses, and is similar to the two local solvers, albeit with larger constant.  The \textit{local random} estimator is much slower than the \textit{local ground truth} estimator and has a larger variance, indicating the dependence on the quality of the initialization. Lastly, we note that the \textit{chordal} estimator has relatively low variance in solution time, which may be desirable for deployment on hardware for applications with strict time budgets.  

The right pane of Figure~\ref{fig:se3-ring-scalability} presents a box plot of the average pose error between the estimate and the ground truth, computed using the log-map norm ($\left\lVert \log\!\left(T_{\mathrm{gt},i}^{-1} T_{i}\right) \right\rVert_2$). The plot shows that the \textit{chordal} estimator has similar accuracy to the \textit{local ground truth} estimator, with the \textit{local random} estimator often resulting in an order-of-magnitude higher average pose error. In addition, \textit{local random} has a large variance in errors, indicating the peril of converging to spurious local minima after poor initialization. Conversely, the \textit{chordal} estimator performs deterministic global optimization and thus has zero variance in average pose error.   

\subsection{2D pose localization example}
\label{sec:2D-loc}
The 2D pose localization example we study has poses in \SE{2}, relative measurements between consecutive poses, and absolute measurements for each pose; the problem can thus be represented by a chain factor graph.

Figure \ref{fig:se2-scalability} shows that, similar to the 3D pose-graph SLAM problem, the proposed \textit{chordal} estimator has a comparable runtime complexity to that of local solvers, but it is even faster thanks to increased sparsity. For certain large estimation problems, the \textit{chordal} estimator is even faster than the \textit{local random} estimator, which is notable since it comes with global optimality guarantees since the convex relaxation is known to be tight. Lastly, we observe that the \textit{chordal} estimator has low variance in solver time and always finds the global minimum, while the \textit{local random} estimator often converges to poor local minima. 




\section{Conclusions and Future Work}
This work demonstrates a scalable certifiable estimator applicable to state estimation problems on arbitrary factor graphs (including those with a ring topology) composed of a subset of commonly used factor and variable types. We present an implementation in \textit{GTSAM} to facilitate use by practitioners. Furthermore, we show that existing \textit{GTSAM} functionality for constructing a Bayes tree automatically yields the chordal decomposition crucial for the estimator. We have shown, for two representative problems, that this chordal estimator has a runtime complexity that empirically is linear in the problem size, similar to local solvers, yet reliably finds certifiably global optima without requiring a high-quality initialization. Future work involves implementing the proposed lifting transformations for a broader range of variable and factor types that arise in different inference problems. A systematic comparison with the alternative structure-exploiting solver based on the Riemannian staircase~\cite{xu2026certifiable}, enabled by its \textit{GTSAM} integration, is also planned. Finally, a promising extension is the implementation of a distributed version of this chordal estimator in \textit{GTSAM}, leveraging splitting methods such as the Alternating Direction Method of Multipliers (ADMM)~\cite{boyd_admm_2010, dumbgen2024exploiting}.

\section*{Acknowledgment}
The work was done in part while FD was with the WILLOW team, Inria Paris, ENS, PSL University and supported by the European Union, through the Horizon Europe research and innovation program under the Marie Skłodowska-Curie (GA no.101207106).


\IEEEtriggeratref{11}

\bibliographystyle{IEEEtran}
\bibliography{zotero-final,refs-fd}

\end{document}